\documentclass[10pt,twocolumn,letterpaper]{article}

\usepackage[pagenumbers]{cvpr} %

\definecolor{cvprblue}{rgb}{0.21,0.49,0.74}
\usepackage[pagebackref,breaklinks,colorlinks,allcolors=cvprblue]{hyperref}

\def\paperID{12927} %
\def\confName{CVPR}
\def\confYear{2025}

\usepackage{times}
\usepackage{epsfig}
\usepackage{graphicx}
\usepackage{amsmath}
\usepackage{amssymb}

\usepackage{url}            %
\usepackage{booktabs}       %
\usepackage{amsfonts}       %
\usepackage{nicefrac}       %
\usepackage{microtype}      %
\usepackage{xcolor}         %

\usepackage{pifont}
\usepackage{color}
\usepackage{wrapfig}
\usepackage{makecell}
\usepackage{colortbl}
\usepackage{multirow}
\usepackage{fixltx2e}
\usepackage{subcaption}
\usepackage{etoolbox}
\usepackage{multicol}
\usepackage{refcount}
\usepackage{enumitem}
\usepackage{comment}
\usepackage[misc]{ifsym}

\newcommand{\cmark}{\ding{51}}%
\newcommand{\xmark}{\ding{55}}%
\definecolor{LightCyan}{rgb}{0.88,1,1}
\definecolor{mygray}{gray}{0.9}
\definecolor{mygray2}{gray}{0.6}

\definecolor{highlight}{HTML}{A3C4FA}

\title{StreamChat: Chatting with Streaming Video}

\author{%
  Jihao Liu$^{1,2,\ast}$ 
  \quad Zhiding Yu$^{2}$~\textsuperscript{\Letter}
  \quad Shiyi Lan$^{2}$ 
  \quad Shihao Wang$^{2,5,\ast}$ 
  \quad Rongyao Fang$^{1}$ \\
  \quad Jan Kautz$^{2}$
  \quad Hongsheng Li$^{1,3,4}$~\textsuperscript{\Letter}
  \quad Jose M. Alvarez$^{2}$ \\
  \small
  $^1$CUHK MMLab \quad
  $^2$NVIDIA \\
  \small
  $^3$Shanghai AI Laboratory \quad
  $^4$CPII under InnoHK \\
  \small
  $^5$The Hong Kong Polytechnic University
}

\begin{document}

\twocolumn[{
\maketitle
\vspace{-3em}
\renewcommand\twocolumn[1][]{#1}
\begin{center}
    \centering
    \includegraphics[width=1.0\textwidth]{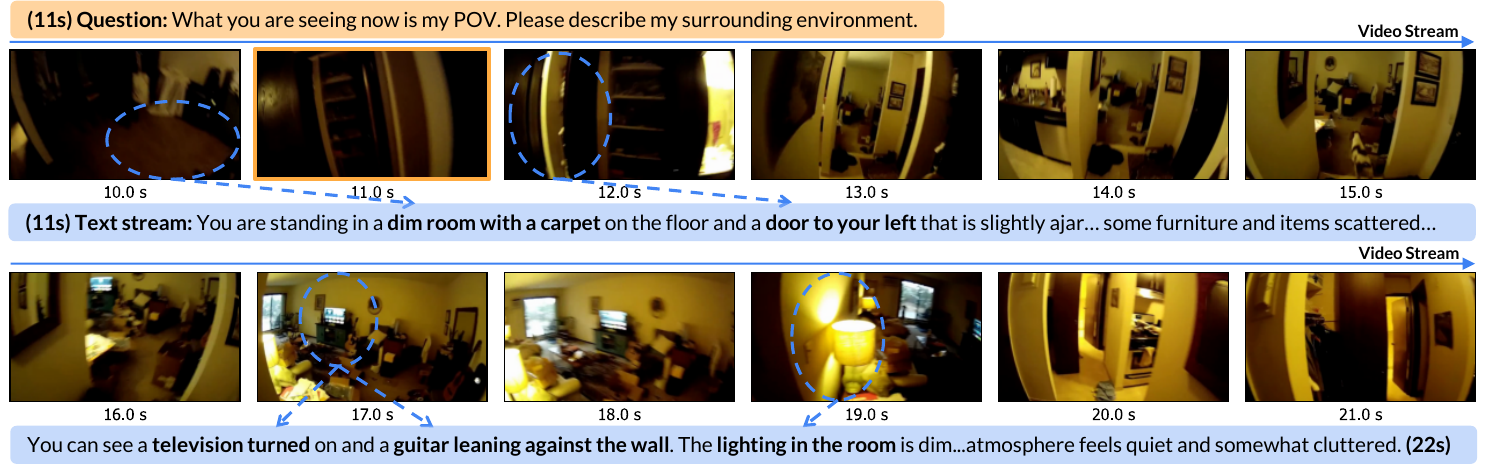}
    \captionof{figure}{\textbf{Example of StreamChat on streaming video.} In the example, the question is asked at the 11th second. As the model outputs its text steam, it continuously follows the dynamic content of the streaming video and uses up-to-date video content to answer the question. }
    \label{figure:teaser}
\end{center}
}]

\renewcommand{\thefootnote}{\fnsymbol{footnote}}
\footnotetext{$\ast$ Work done during an internship at NVIDIA}
\footnotetext{\Letter \ Corresponding authors}

\begin{abstract}

This paper presents StreamChat, a novel approach that enhances the interaction capabilities of Large Multimodal Models (LMMs) with streaming video content. In streaming interaction scenarios, existing methods rely solely on visual information available at the moment a question is posed, resulting in significant delays as the model remains unaware of subsequent changes in the streaming video. StreamChat addresses this limitation by innovatively updating the visual context at each decoding step, ensuring that the model utilizes up-to-date video content throughout the decoding process. Additionally, we introduce a flexible and efficient cross-attention-based architecture to process dynamic streaming inputs while maintaining inference efficiency for streaming interactions. Furthermore, we construct a new dense instruction dataset to facilitate the training of streaming interaction models, complemented by a parallel 3D-RoPE mechanism that encodes the relative temporal information of visual and text tokens.
Experimental results demonstrate that StreamChat achieves competitive performance on established image and video benchmarks and exhibits superior capabilities in streaming interaction scenarios compared to state-of-the-art video LMMs. Our project page is at \href{https://jihaonew.github.io/projects/streamchat.html}{StreamChat}.
\end{abstract}

\section{Introduction}
\label{sec:intro}
The recent surge of large language models (LLMs)~\cite{kenton2019bert,raffel2020exploring,mann2020language,bai2023qwen,touvron2023llama,llama3} and large multimodal models (LMMs)~\cite{qwen2vl,llava_ov,internvl,cambrian} has unlocked numerous application scenarios, including visual instruction following~\cite{llava,llava_1_5,llava_next,mm_instruct} and long video understanding~\cite{longva,longvila}. 
Notably, frontier models such as GPT-4o~\cite{openai2024gpt4o} and Gemini~\cite{team2023gemini} have shown remarkable proficiency when interacting with streaming videos, attracting considerable interest in the field. While recent open approaches~\cite{videollm_online,flash_vstream,videollm_mod,mini_omni2} have emerged to enhance streaming video processing, they still fall short in interaction fluency and perceptual capabilities.

\begin{figure}
    \centering
    \includegraphics[width=1.0\linewidth]{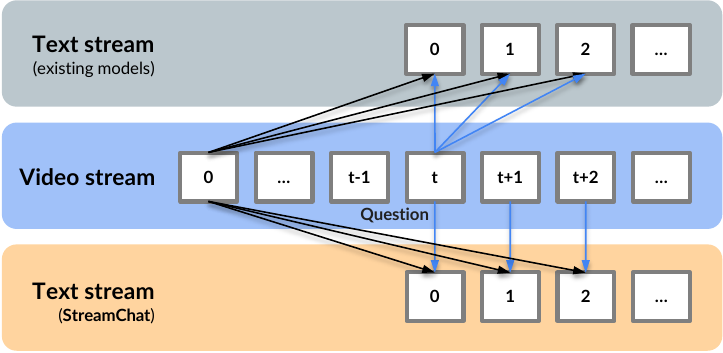}
    \caption{\textbf{Comparison of context in the decoding process with existing models.} For each text token, the black and blue arrows indicate the beginning and end of the utilized visual context, respectively. While existing models (top) use a fixed visual context when decoding, StreamChat (bottom) aligns the video and text streams temporally and dynamically updates its visual context based on the streaming video.}
    \label{fig:comparison}
\end{figure}

To enable effective interaction with streaming videos, LMMs must not only accurately identify the visual content of each frame but also track dynamic changes in the streaming video, leveraging the latest visual information to answer questions, as illustrated in~\Cref{figure:teaser}.
Despite notable progress in video understanding of LMMs~\cite{qwen2.5,llava_video,videollama,videollama2,pllava}, existing models often overlook the crucial need to capture dynamic changes, negatively impacting the interaction experience. 
Specifically, current methods typically rely on video information only up to the moment a question is asked; however, the streaming content may change significantly during the decoding process, leaving the model unaware of these updates. For instance, assume a question is posed at time $t$ and the model takes $t'$ seconds to answer the question, existing methods only utilize the video content from the interval \(0\) to \(t\) to answer the question, leaving the model unaware of any changes that occur between \(t\) and \(t+t'\).
This \emph{delay} can be particularly detrimental in highly dynamic video environments or when the answer to a question is lengthy, resulting in a suboptimal user experience. We illustrate the problem in~\Cref{fig:comparison} (top).

To address these limitations, we propose StreamChat, a novel approach that enables LLMs to interact dynamically with streaming video content. The core idea is to provide the LLM with the latest video information at each decoding step, allowing it to better capture video dynamics and adjust its responses accordingly, which is illustrated in~\Cref{fig:comparison} (bottom).
Mechanistically, StreamChat enhances the model's ability to interact with streaming video data, ensuring more temporally aligned responses, as demonstrated in~\Cref{figure:teaser}.
To effectively handle the dynamic visual inputs of streaming videos, we design a flexible and efficient architecture based on cross-attention mechanism~\cite{attention,flamingo,llama3}, bridging the LLM and visual inputs in StreamChat. The cross-attention design facilitates processing variable-length inputs in the streaming scenario and is more efficient when dealing with a large number of visual tokens.

To facilitate the training of streaming interaction models, we introduce a dense instruction dataset to train StreamChat. Existing video instruction-tuning datasets~\cite{llava_video,mvbench,coin,densecap_anet,ego4d} primarily focus on offline video understanding, \ie, the model can perceive the complete video before answering a question, which is different from the streaming interaction scenarios where the video content is dynamically changing during the answering process. To bridge this gap, we create a new dense instruction dataset based on existing dense caption datasets. One dense instruction data consists of several (time interval, instruction, answer) triplets, with each word of the instruction-answer pairs annotated with a timestamp in a heuristic manner. During training, we employ attention masks to ensure that each text token can only attend to video information before its corresponding timestamp. 
This method effectively simulates the conditions of streaming interaction throughout the training process.

Importantly, we do not directly input the absolute timestamp of each token into the model, as these timestamps are unavailable during inference. Instead, we propose a parallel 3D-RoPE mechanism that allows each token to be aware of its relative temporal position within the video. We use three components in RoPE~\cite{rope} to represent temporal, height, and width, respectively. Unlike existing approaches that arrange video and text in an interleaved manner~\cite{rope,qwen2vl}, 
our method organizes them in a parallel way to ensure that visual and text tokens at the same timestamp share the same temporal context in RoPE, which enhances the continuity during streaming interactions.

Through extensive experiments, we demonstrate that StreamChat not only achieves competitive performance on established image and video benchmarks but also significantly improves capabilities in streaming interaction scenarios. Specifically, we create a benchmark designed to evaluate LMMs in streaming interaction scenarios. We demonstrate that our StreamChat-7B outperforms the state-of-the-art LLaVA-Video-72B model.

\section{Methods}
\label{sec:methods}
Recent advancements in large multimodal models (LMMs)~\cite{qwen2.5,llava_video,videollama,videollama2,pllava} have significantly improved the models' video understanding capabilities. However, in streaming interaction scenarios, LMMs must also accurately capture the dynamic changes of the streaming video content, which is overlooked by existing models.
To bridge this gap, we propose StreamChat, a novel LMM that can interact smoothly with streaming videos and track the latest changes in the videos to refine its answers. In this section, we outline the methodology underlying StreamChat, detailing the architectural innovations and techniques employed to enable dynamic interaction with streaming video. We begin by describing the StreamChat architecture design in~\Cref{sec:methods:arch} and then introduce how we generate and construct our training data in~\Cref{sec:methods:data}. Finally, we discuss the development of our training and inference pipeline in~\Cref{sec:methods:training}.

\begin{figure}
    \centering
    \includegraphics[width=0.8\linewidth]{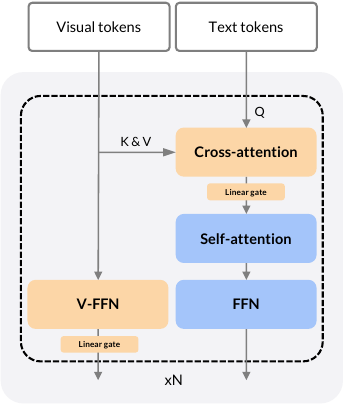}
    \caption{\textbf{The StreamChat architecture.} We utilize cross-attention blocks to bridge the visual and text tokens and V-FFN blocks to update the visual tokens throughout the LLM's forward process. Those two blocks' outputs are scaled with a linear gate mechanism.}
    \label{fig:architecture}
\end{figure}

\subsection{StreamChat Architecture}
\label{sec:methods:arch}

To support streaming video content, we design a flexible and efficient architecture capable of handling dynamic video inputs through the cross-attention mechanism. Additionally, we introduce visual feedforward network (V-FFN) experts to enhance the visual representations 
throughout the large language model's (LLM) forward process. We also propose a parallel 3D-RoPE mechanism to better encode the temporal information in streaming interaction scenarios. 
The architecture is illustrated in~\Cref{fig:architecture}.

\paragraph{Cross attention.}
We build a cross-attention-based architecture to bridge the visual and text tokens. 
Given an input streaming video, we utilize a pretrained vision model to extract visual tokens for each sampled frame separately. To integrate these visual tokens with the LLM, we insert several cross-attention blocks into the LLM architecture, where text tokens serve as queries and visual tokens act as keys and values. The visual tokens are dynamically updated during the interaction process, and the cross-attention design facilitates processing these dynamic inputs. Moreover, compared to self-attention-based architectures (\eg, LLaVA~\cite{llava}), cross-attention is significantly more efficient when the visual tokens are much more than the text tokens, especially in streaming interactions, where we have high frames per second (FPS) for inference. In practice, our cross-attention blocks share parameters with the self-attention blocks of the LLM to improve the convergence speed during training.

We further utilize V-FFN experts to enhance the visual representations 
throughout the LLM's forward process. 
Specifically, after each cross-attention block, we update the visual tokens with a V-FFN expert and feed the updated tokens into the subsequent cross-attention block. 
In contrast to previous cross-attention-based models~\cite{flamingo,evlm,mplug_3,llama3} that utilize the same visual representations for all cross-attention blocks, our V-FFN experts allow the visual representations to better align with the LLM's hidden status and improve the final performance.
Practically, these V-FFN experts are initialized from the LLM's FFN instead of training from scratch to inherit the pretrained knowledge of the LLM.

Previous cross-attention-based models~\cite{flamingo,llama3} typically employ a tanh-gating mechanism to ensure that the language model produces the same results as the original LLM at the early stage and stabilize the training. However, the tanh function suffers from the gradient vanishing problem, which results in suboptimal performance. Instead, we introduce a linear gate to scale the output of cross-attention and V-FFN blocks to a relatively small range during the initial training phase following CaiT~\cite{layer_scale}. 
The linear gate mechanism mitigates the gradient problem while also stabilizing the training process.

\begin{figure}
    \centering
    \includegraphics[width=1.0\linewidth]{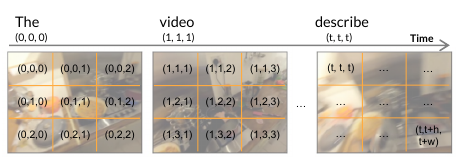}
    \caption{\textbf{The parallel 3D-RoPE}. For visual and text tokens at the same timestamp, they share the same temporal position.} 
    \label{fig:rope}
\end{figure}

\paragraph{Parallel 3D-RoPE.}
To better model the positional information of streaming video and text, we propose parallel 3D-RoPE that extends traditional 1D-RoPE~\cite{rope} to 3D space with a parallel arrangement of visual and text tokens. Specifically, we split the embedding of RoPE into three components.
For text tokens, these components are identical to represent the temporal location of each token. For visual tokens, these three components represent the temporal, height, and width locations of each token. Unlike previous approaches that arrange the visual and text tokens in an interleaved manner~\cite{qwen2vl}, we use a parallel way to arrange them, as illustrated in~\Cref{fig:rope}. Given a text token and a visual token at the same timestamp, we apply the same temporal index for them. Our intuition is that in the streaming setting, one specific timestamp's text and visual tokens are happening simultaneously, and therefore should share the same temporal location instead of an interleaved location. The parallel arrangement is crucial for the high FPS inference in the streaming setting, where the traditional arrangement may have a significant temporal positional gap between two adjacent text tokens while our approach ensures their continuity.

\subsection{Dense Instruction Data}
\label{sec:methods:data}
Existing video instruction-tuning datasets~\cite{llava_video,mvbench,coin,densecap_anet,ego4d} have made significant progress in offline video understanding, \ie, the model can see the whole video before answering a question. 
However, these datasets are not suitable for training streaming interaction models, where the input is a streaming video and each text token can only see part of the video. For instance, a text token at timestamp $t$ can only perceive video frames at and before timestamp $t$. To tackle this problem, we create a new video instruction-tuning dataset from the existing dense caption datasets, whose captions are paired with timestamp intervals. 

Given a video with its dense caption, we prompt an LLM (\eg, Gemini-1.5-Pro~\cite{team2023gemini}) to pick a start time of one video segment and then generate an instruction-answer pair based on the caption of this segment. We instruct the LLM to focus on streaming interaction scenarios and generate relevant instructions. To enhance the diversity of the instruction data, we initially generate 5k pairs and conduct a clustering to eliminate highly similar instructions. We manually review the remaining examples and use them as in-context examples for subsequent data generation. Ultimately, we collect a total of 51k examples from two dense caption datasets, Ego4D~\cite{ego4d} and Vript~\cite{vript}, with one representing the ego-centric environment and the other one representing the natural environment.

\subsection{Training and Inference}
\label{sec:methods:training}

\paragraph{Data arrangement.}
Given the initial instruction data with coarse temporal annotations, \ie, in the form of (time interval, instruction, answer), we employ a heuristic approach to assign timestamps to each word in the instruction data.
For instance, consider a triplet where the time interval is 5-10 seconds, the instruction is ``What is the person in the video doing now?" and the answer is ``The person is cooking right now." To generate fine-grained temporal annotations, we transform this coarse-grained triplet into a sequence of words including time indicators. The transformation results in the following format:
\begin{align*}
\begin{array}{l}
    \texttt{\scriptsize \textbf{Instruction:}<5>What is the person in the video doing now?} \\
    \texttt{\scriptsize \textbf{Answer:}<5>The <6>person <7>is <8>cooking <9>right <10>now.}
\end{array}
\end{align*}

Here, $\texttt{<t>}$ represents the $t$-th second. The intuition behind this design is that the instruction is input by the user instantaneously, while the answer is decoded token by token by the model. For this example, we assume the model decodes one token per second. Note that the $\texttt{<t>}$ indicators are not directly input to the model but serve solely for reference.

\paragraph{Attention mask.} To ensure that the token at time $\texttt{<t>}$ does not attend to video frames occurring after $\texttt{<t>}$, we utilize attention mask to block such attention. This mechanism is crucial for maintaining the temporal integrity of the streaming interaction, allowing the model to focus only on relevant visual information available at each decoding step.

\paragraph{Inference.}
During inference, StreamChat employs a parallel approach to ensure efficient processing of the streaming video content. Specifically, we utilize a separate thread to continuously read the video stream and store the extracted visual tokens in a First-In-First-Out (FIFO) queue. When the LLM requires decoding to generate a response, it acquires the latest video tokens from the FIFO queue. 
The model then incorporates this current information to decode the next token, ensuring that its responses are informed by the most up-to-date video stream context.
This design not only enhances the relevance of the model's outputs but also supports seamless streaming interactions, allowing users to engage with dynamic video content effectively.

\section{Experiment Setups}
\label{sec:exp_setup}

In this section, we outline the experimental setups employed in our study. We build our model using the SigLIP vision encoder~\cite{siglip} with PaliGemma weights~\cite{paligemma} and 7B/14B Qwen 2.5 LLM~\cite{qwen2.5}. We utilize a Multi-Layer Perceptron (MLP) adapter~\cite{llava} to align the hidden dimensions of the vision and language components.

\begin{figure*}
    \centering
    \includegraphics[width=1.0\linewidth]{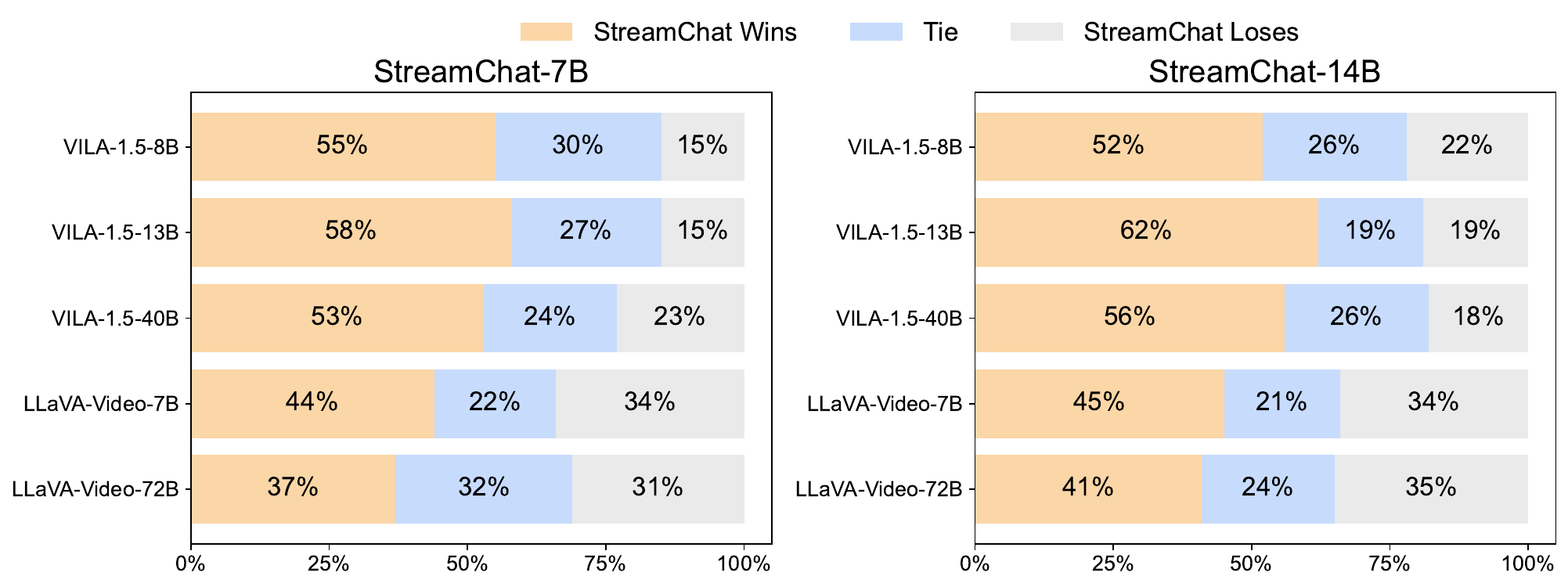}
    \vspace{-1em}
    \caption{\textbf{Comparison of StreamChat with leading video LMMs on streaming evaluation.} We use StreamChat-7B/-14B as one of the candidate models and report the win/tie/loss rate against VILA or LLaVA-Video models. Our StreamChat models demonstrate stronger streaming interaction capabilities, and can even outperform LLaVA-Video-72B which uses a much larger base LLM.} 
    \label{fig:win_rate}
\end{figure*}

\begin{figure*}
    \centering
    \includegraphics[width=1.0\linewidth]{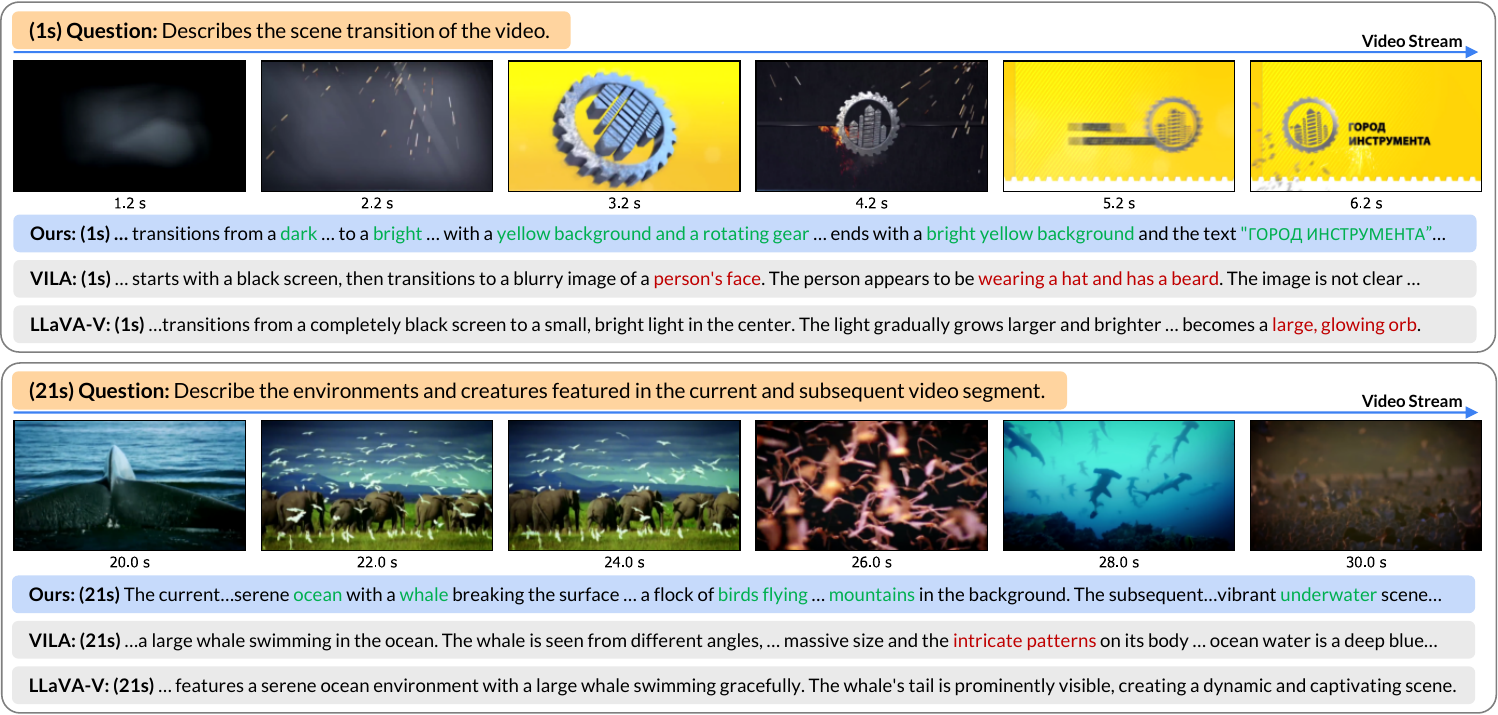}
    \vspace{-1em}
    \caption{\textbf{Qualitative evaluation of StreamChat on streaming video.} In the example shown, the questions are asked at the first second (top) and the 21st second (bottom), respectively.
    Our model can capture the dynamic video content and adapt its answer accordingly. In comparison, VILA and LLaVA-Video fail to follow the streaming video and exhibit factual errors (highlighted in red).}
    \label{fig:example}
\end{figure*}

\subsection{Pretraining}
We implement a two-stage pretraining process and gradually unfreeze the pretrained parameters for more effective pretraining. In both stages, we utilize a combination of ReCap data from LLaVA-Next~\cite{llava_next}, part of InternVL pretraining data~\cite{internvl}, MMC4~\cite{mmc4}, and dense caption datasets~\cite{ego4d,youcook2,densecap_anet,coin,vript}. In stage 1, we only train the MLP adapter for alignment. We train the MLP for 5000 steps with a maximum learning rate of $5 \times 10^{-4}$ and batch size of 512. In stage 2, we further unfreeze the vision encoder and the visual feedforward network (V-FFN) experts to achieve deeper alignment. We train for 5000 steps with a maximum learning rate of $2 \times 10^{-5}$ and batch size of 512. For the dense caption datasets, we employ 1 frame per second (FPS) and a maximum of 40 frames for training. For other video data, we uniformly sample 40 frames for training. In total, we utilize 5.1 million samples for pretraining.

\subsection{Instruction Tuning}
We construct a comprehensive instruction-tuning dataset mainly based on Eagle-1.8M~\cite{eagle}.
In addition, we incorporate our dense instruction dataset and LLaVA-Video~\cite{llava_video} for instruction tuning. 
We unfreeze all the parameters and train on the dataset combination for 1 epoch. We use a maximum learning rate of \(2 \times 10^{-5}\) and a batch size of 768. For the dense instruction data, we use 1 FPS and a maximum of 32 frames for training. For other video instruction data, we uniformly sample 32 frames for training. In total, we use 2.9 million samples for instruction tuning.

\begin{table*}[t]
    \centering
    \setlength{\tabcolsep}{4pt}
    \begin{tabular}{rc | cccc cccc cc}
    Method &
    \rotatebox{90}{\# Vis Tok.} &
    \rotatebox{90}{MME$^\text{P}$} &
    \rotatebox{90}{MMB} &
    \rotatebox{90}{MMMU$^\text{V}$} &
    \rotatebox{90}{MMStar$^\text{V}$} &
    \rotatebox{90}{SEED$^\text{I}$} &
    \rotatebox{90}{GQA} &
    \rotatebox{90}{SQA$^\text{I}$} &
    \rotatebox{90}{AI2D} &
    \rotatebox{90}{TextVQA}  &
    \rotatebox{90}{RealworldQA} \\
    \hline
    \rowcolor{gray!10}
    \multicolumn{2}{l|}{\textit{Private models}} & \multicolumn{10}{l}{} \\
    GPT-4V  & UNK. & 1409 & 75.8 & 56.8 & 57.1 & 69.1 & 36.8 & 75.7 & 78.2 & 78.0 & 61.4 \\
    Gemini-1.0 Pro & UNK. & 1496 & 73.6 & 47.9 & 42.6 & 70.7 & - & 79.5 & - & - & - \\
    Gemini-1.5 Pro & UNK. & - & - & 58.5 & - & - & - & - & 80.3 & 73.5 & 67.5 \\
    Grok-1.5 & UNK.& - & - & 53.6 & - & - & - & - & 88.3 & 78.1 & 68.7  \\
    
    \rowcolor{gray!10}
    \multicolumn{2}{l|}{\textit{7B-Level Base LLM}} & \multicolumn{10}{l}{} \\
    Mini-Gemini-HD-8B & 2880 & \textbf{1606} & 72.7 & 37.3 & - & 73.2 & 64.5 & 75.1 & 73.5 & 70.2 & 62.1 \\
    LLaVA-NeXT-8B & 2880 & 1603 & 72.1 & 41.7 & - & 72.7 & \textbf{65.2} & 72.8 & 71.6 & 64.6 & 60.1 \\
    Cambrian-1-8B  & 576 & 1547 & \textbf{75.9} & 42.7 & - & \textbf{74.7} & 64.6 & 80.4 & 73.0 & 71.7 & \textbf{64.2} \\
    \rowcolor{green!10} StreamChat-7B  & \textbf{256} & 1520 & 74.4 & \textbf{48.1} & \textbf{46.0} & 74.3 & 62.4 & \textbf{85.5} & \textbf{76.6} & \textbf{72.4} & 61.7 \\
    
    \rowcolor{gray!10}
    \multicolumn{2}{l|}{\textit{14B-Level Base LLM}} & \multicolumn{10}{l}{} \\
    Mini-Gemini-HD-13B & 2880 & 1597 & 68.6 & 37.3 & - & 70.6 & 63.7 & 71.9 & 70.1 & 70.2 & 57.5 \\
    LLaVA-NeXT-13B & 2880 & 1575 & 70.0 & 36.2 & - & 65.6 & \textbf{65.4} & 73.5 & 70.0 & 67.1 & 59.1 \\
    Cambrian-1-13B  & 576 & 1610 & 75.7 & 40.0 & - & 74.4 & 64.3 & 79.3 & 73.6 & 72.8 & 63.0 \\
    \rowcolor{green!10} StreamChat-14B  & \textbf{256} & \textbf{1617} & \textbf{79.0} & \textbf{50.1} & \textbf{53.6} & \textbf{75.5} & 63.3 & \textbf{85.8} & \textbf{79.5} & \textbf{74.4} & \textbf{63.3} \\
    \end{tabular}
\caption{\textbf{Comparison of StreamChat with leading LMMs on image benchmarks.} StreamChat achieves competitive performance on these benchmarks while using only 256 visual tokens. }
\label{tab:results_image}
\end{table*}

\section{Streaming Evaluation}
\label{sec:qualitative}
To evaluate large multimodal models' (LMMs) streaming interaction capabilities, we construct a streaming evaluation benchmark from existing dense caption datasets. Based on the dense caption of a video, we prompt Gemini-1.5-Pro to generate an instruction-answer pair for a specific timestamp. We remove samples that are not related to streaming scenarios and manually review and refine each remaining sample to ensure that the instruction and answer align with the video content and timestamp. Ultimately, we collect 100 evaluation samples, with 80 sourced from Vript~\cite{vript} and 20 from Ego4D~\cite{ego4d}.

Following~\cite{mm_instruct,llava}, we use Gemini-1.5-Pro as the judge for performance evaluation. 
Given a video and its corresponding instruction, we infer two candidate models to predict their respective answers. We then feed the ground truth answer along with the outputs from the two models to the judge. The judge is required to evaluate the two answers on adherence, helpfulness, relevance, and accuracy. We prompt the judge to determine which model's answer is better or if both are tied in terms of quality and then require it to provide a detailed justification explaining its reasoning based on the judgment. We use our StreamChat model as one of the candidate models and calculate the overall win rate in comparison to the other models.

\subsection{Quantitative Results}

We show the comparison between StreamChat and other video LLMs in~\Cref{fig:win_rate}. The frames per second (FPS) is set to 5. We use 32 frames for StreamChat and LLaVA-Video~\cite{llava_video} models, and 16 frames for VILA~\cite{vila} since using 32 would exceed its context range. Our results demonstrate that our StreamChat models exhibit superior streaming interaction capabilities compared to LLaVA-Video and VILA models. Notably, compared to VILA-1.5-40B, our StreamChat-7B model produces equally or more preferable answers in 77\% of the evaluation cases despite using a much smaller LLM. While LLaVA-Video models excel in offline video understanding, StreamChat-7B outperforms them in streaming interaction scenarios, highlighting the importance of capturing video dynamics during streaming inference. Furthermore, we observe that our StreamChat-14B demonstrates overall better performance than StreamChat-7B, indicating that scaling the base LLM can also improve the streaming interaction performance. 

\begin{table*}[t]
\setlength{\tabcolsep}{4pt}
\centering
\begin{tabular}{rc|cccc cccc}
    Method &
    \rotatebox{90}{\# Frames} &
    \rotatebox{90}{ActNet-QA} &
    \rotatebox{90}{EgoSchema} &
    \rotatebox{90}{MLVU} &
    \rotatebox{90}{MVBench} &
    \rotatebox{90}{NExT-QA} &
    \rotatebox{90}{PerceptionTest} &
    \rotatebox{90}{LongVideoBench} &
    \rotatebox{90}{VideoMME} \\
    \hline

    \rowcolor{gray!10}
    \multicolumn{2}{l|}{\textit{Private models}} & \multicolumn{8}{l}{} \\
    GPT-4V & UNK. & 57.0 & - & 49.2 & 43.5 & - & -   & 61.3 & 59.9/63.3 \\
    GPT-4o & UNK. & - & - & 64.6 & - & - & -    & 66.7 & 71.9/77.2 \\
    Gemini-1.5-Flash & UNK. & 55.3 & 65.7 & - & -  & - & -   & 61.6  & 70.3/75.0 \\
    Gemini-1.5-Pro & UNK. & 57.5 & 72.2 & - & -  & - & -  & 64.0 & 75.0/81.3 \\ 
    
    \rowcolor{gray!10}
    \multicolumn{2}{l|}{\textit{7B-Level Base LLM}} & \multicolumn{8}{l}{} \\
    LongVA-7B & 128 & 50.0 & - & 56.3 & - & 68.3 & -  & -  & 52.6/54.3 \\
    IXC-2.5-7B & 64 & 52.8 & - & 37.3 & \textbf{69.1} & 71.0 & 34.4    & - & 55.8/58.8 \\   
    PLLaVA-7B & 16 & \textbf{56.3} & - & - & 46.6  & - & - & 40.2 & - \\
    VideoLLaMA2-7B & 16 & 50.2 & \textbf{51.7} & 48.5 & 54.6 & - & -  & - & 47.9/50.3 \\ 
    \rowcolor{green!10} StreamChat-7B & 40 & 54.9 & 48.4 & \textbf{63.9} & 53.3 & \textbf{78.5} & \textbf{63.0} & \textbf{54.2} & \textbf{58.6/62.8} \\ 
    
    \rowcolor{gray!10}
    \multicolumn{2}{l|}{\textit{14B+ Level Base LLM}} & \multicolumn{8}{l}{} \\
    VILA-40B & UNK. & 58.0 & 58.0 & - & - & 67.9 & 54.0  & -  & 60.1/61.1 \\
    PLLaVA-13B & 16 & 56.3 & - & - & 50.1  & - & - & 45.6 & - \\
    PLLaVA-34B & 16 & \textbf{60.9} & - & - & 58.1  & - & - & 53.2 & - \\
    VideoLLaMA2-72B & 16 & 55.2 & \textbf{63.9} & 61.2 & \textbf{62.0} & - & -  & - & 61.4/63.1 \\ 
    
    \rowcolor{green!10} StreamChat-14B & 40 & 55.9 & 57.2 & \textbf{66.6} & 55.2 & \textbf{79.4} & \textbf{63.7} & \textbf{57.1} & \textbf{63.1/66.3} \\ 
    \end{tabular}%
\caption{\textbf{Comparison of StreamChat with leading LMMs on video benchmarks.} StreamChat achieves competitive performance on these benchmarks and even outperforms models with a much larger base LLM.
StreamChat's cross-attention-based architecture is efficient in processing a large number of video frames.}
\label{tab:results_video}
\end{table*}

\subsection{Qualitative Results}

We provide a qualitative evaluation of StreamChat's capabilities on streaming video, as illustrated in~\Cref{fig:example}. In the example, we pose a question at a specific timestamp. While previous methods only answer the question using the visual context up to the moment the question is asked, StreamChat can dynamically update its visual context alongside the streaming video and adapt its answer accordingly. We show that StreamChat can better capture dynamic video content and provide more accurate answers. In contrast, VILA~\cite{vila} and LLaVA-Video~\cite{llava_video} struggle to maintain temporal alignment with the streaming video and exhibit factual errors (highlighted in red).

\section{Benchmark Results}
\label{sec:main_results}

We evaluate the performance of StreamChat models on popular image~\cite{fu2023mme,liu2023mmbench,mmmu,mmstart,seed,hudson2019gqa,sqa,ai2d,singh2019textvqa,realworldqa} and video benchmarks~\cite{activitynetqa,egoschema,mlvu,mvbench,nextqa,perception_test,longvideobench,videomme} using the LMMs-Eval library~\cite{lmms}. Note that to maintain StreamChat's efficiency for streaming interactions, we do not employ multiple vision encoders~\cite{eagle,cambrian} or image tiling techniques~\cite{llava_next,internvl}, which could compromise performance on benchmarks requiring high-resolution inputs.

The performance of StreamChat on image benchmarks is presented in ~\Cref{tab:results_image}. StreamChat demonstrates strong results compared to Cambrian-1, which utilizes multiple vision encoders, and LLaVA-NeXT, which employs image tiling. Notably, our StreamChat-7B achieves a score of 48.1 on the MMMU benchmark, surpassing LLaVA-NeXT-8B and Cambrian-1-8B by 6.4 and 5.4 points, respectively. Additionally, StreamChat outperforms both LLaVA-NeXT and Cambrian-1 on TextVQA, despite using significantly fewer visual tokens. Overall, StreamChat achieves competitive performance on image benchmarks while ensuring computation efficiency.

We present StreamChat's performance on video benchmarks in~\Cref{tab:results_video}. Our model significantly outperforms PLLaVA~\cite{pllava} and VideoLLaMA2~\cite{videollama2}, using a 7B-level base LLM. Specifically, we achieve scores of 58.6/62.8 on the VideoMME benchmark~\cite{videomme}, outperforming VideoLLaMA2-7B by 10.7/11.5 points. Moreover, StreamChat-14B demonstrates superior performance compared to VILA-40B and VideoLLaMA2-72B, which utilize much larger base LLMs. Importantly, our model remains efficient even when processing more frames for inference, as our cross-attention-based architecture mitigates the heavy computation associated with self-attention across frames.

\begin{table*}[t]
    \centering
    \setlength{\tabcolsep}{4pt}
    \begin{tabular}{cccc | cccc| cccc| c}
    V-FFN &
    \makecell{Linear \\ Gate} &
    \makecell{Param. \\ Reuse} &
    \makecell{Dense \\ Instruction} &
    \rotatebox{90}{MMMU$^\text{V}$} &
    \rotatebox{90}{AI2D} &
    \rotatebox{90}{TextVQA}  &
    \rotatebox{90}{RealworldQA} & 
    \rotatebox{90}{MLVU} &
    \rotatebox{90}{MVBench} &
    \rotatebox{90}{PerceptionTest} &
    \rotatebox{90}{VideoMME} & 
    \makecell{StreamEval \\ Win/Tie/Loss} \\
    \hline
    
    \xmark & \cmark & \cmark & \cmark & \textbf{46.7} & 75.7 & 62.7 & 57.8 & 58.2 & 47.3 & 49.3 & 51.1 & 18/46/36 \\
    \cmark & \xmark & \cmark & \cmark & 45.1 & 74.8 & 60.7 & 58.3 & \textbf{60.0} & 49.4 & 51.8 & 52.4 & 25/45/30 \\
    \cmark & \cmark & \xmark & \cmark & 44.4 & 72.4 & 46.9 & 46.3 & 53.5 & 43.4 & 46.6 & 47.1 & 20/33/47 \\
    \cmark & \cmark & \cmark & \xmark & 46.0 & \textbf{76.5} & 62.5 & \textbf{59.4} & 57.0 & 49.0 & \textbf{52.8} & 51.3 & 25/34/41 \\
    \rowcolor{green!10}
    \cmark & \cmark & \cmark & \cmark & 45.2 & 76.1 & \textbf{63.3} & 58.0 & 59.7 & \textbf{49.5} & 51.1 & \textbf{52.6} & -/-/- \\
    \end{tabular}
\vspace{-1em}
\caption{\textbf{Ablation study results.} StreamEval indicates our proposed streaming evaluation, in which we use our final solution (last row) as one of the candidate models and report other models' performance against our final solution.}
\label{tab:ablation}
\end{table*}

\section{Ablation Studies}
\label{sec:ablation}

We use a relatively efficient setting for ablation studies. We shrink the total training steps to 2000 while keeping other hyperparameters the same as our full training. In the instruction tuning stage, unless otherwise specified, we employ a combination of our dense instruction dataset and Eagle-1.8M~\cite{eagle} and train on the combination for 1 epoch. The training hyperparameters are the same as our full training.

We present the results of our ablation study in~\Cref{tab:ablation}, where we ablate our architectural designs and the proposed dense instruction dataset. We compare performance across four image benchmarks, four video benchmarks, and our streaming evaluation. In the streaming evaluation, we employ our StreamChat solution (last row) as one of the candidate models and report the performance of other models relative to StreamChat.

Our experiment results indicate that the architectural enhancements we introduced lead to improved overall performance. Specifically, the StreamChat model outperforms the version without the visual feedforward network (V-FFN) experts on 8 out of 9 benchmarks. Additionally, we observe that using a tanh gate facilitates faster convergence during the early stages of training; however, it ultimately results in poorer final performance compared to the linear gate. The linear gate improves performance on 6 out of 9 benchmarks when compared to the tanh gate. Furthermore, we observe a significant training instability when not reusing the LLM's parameter, which also leads to poor final performance. Our final solution substantially outperforms the model without parameter reuse across all evaluated benchmarks.

When compared to the model trained without dense instruction data, our final solution performs comparably on existing image and video benchmarks. However, in the streaming evaluation, we demonstrate that training on our dense instruction dataset significantly enhances interaction capabilities. Our final solution produces equally or more preferable answers in 75\% of the evaluation cases, indicating that reliance on existing image or video instruction tuning datasets alone is insufficient for effective streaming interactions.

\section{Related Works}
\label{sec:related_works}
\noindent\textbf{Large Multimodal Models.} 
Large multimodal models (LMMs) have garnered significant attention for their robust zero-shot capabilities across various tasks, including image captioning~\cite{coco} and visual question answering (VQA)~\cite{goyal2017vqav2,marino2019okvqa,gurari2018vizwiz}. Notably, Flamingo~\cite{alayrac2022flamingo} showcases visual in-context learning by training on extensive interleaved image-text datasets. GPT-4V~\cite{gpt4} exhibits emerging image-understanding capabilities, providing coherent responses to user queries. In the open-source domain, LLaVA reproduces aspects of GPT-4V's functionality by fine-tuning on generated instruction-following data. Subsequent works, including LLaVA-1.5~\cite{llava_1_5}, Qwen-VL~\cite{bai2023qwenvl}, and CogVLM~\cite{wang2023cogvlm}, aim to enhance model capabilities through architectural refinements, improved training methodologies, and higher-quality training datasets. More recently, open models like InternVL2~\cite{internvl}, LLaVA-OneVision~\cite{llava_ov}, and Qwen2-VL~\cite{qwen2vl} have demonstrated even better performance than state-of-the-art closed models like GPT-4o~\cite{openai2024gpt4o} or Gemini-1.5-Pro~\cite{team2023gemini}, paving the path for research of LMMs.

\noindent\textbf{Multimodal Video Models.} More recent LMMs also incorporate video understanding capability to support a wider range of application scenarios~\cite{li2023videochat,qwen2vl,pllava,videollama,vila,yao2024minicpm}. To support extreme long video understanding, various techniques are proposed to process more video frames during training and inference~\cite{longva,moviechat_plus,song2024moviechat,longvila}.
Additionally, to reduce the redundancy of video information, more recent approaches also propose to compress the video via spatial or temporal compression~\cite{weng2024longvlm,llama_vid,videollamb,nvila}. 
However, these methods focus on offline video understanding, where the entire video is available beforehand. In contrast, StreamChat focuses on processing streaming video, which is more suitable for real-world applications and interaction.

\noindent\textbf{Streaming Video Models.}
The advent of streaming video models began with OpenAI's GPT-4o~\cite{openai2024gpt4o}, which has demonstrated remarkable proficiency in real-time interaction with streaming videos, attracting considerable interest in the field. Following this landmark development, several subsequent works have aimed to enhance large multimodal models for processing streaming video content. Notable approaches include methods such as VideoLLM-online~\cite{videollm_online}, Flash-VStream~\cite{flash_vstream}, and subsequent methods~\cite{fu2024vita,wang2024videollm,videollm_mod,qian2024streaming,mini_omni,mini_omni2,lin2024streamingbench,internlm_xcom_omni}
, which focus on improving the fluency or responsiveness of models during streaming video processing. However, unlike these existing models, which often rely on fixed video content up to the moment a question is posed to answer questions, our work emphasizes the dynamic updating of the visual context during the decoding process, thereby significantly enhancing the interactive experience and mitigating the detrimental delays inherent in highly dynamic environments.

\section{Conclusion}
\label{sec:conclusion}
This paper presents StreamChat, a novel approach that enhances the real-time interaction capabilities of large multimodal models (LMMs) with streaming video content. StreamChat is built on a flexible and efficient cross-attention-based architecture with visual feedforward network (V-FFN) experts. By continuously updating the visual context at each decoding step, StreamChat effectively captures the dynamic changes of streaming video content, leading to temporally aligned responses. We also introduce a dense instruction dataset to facilitate the training of streaming interaction models, alongside a parallel 3D-RoPE mechanism to better arrange the streaming video and text. Our extensive evaluations on both established image and video benchmarks and a novel streaming benchmark demonstrate that StreamChat not only achieves competitive performance on existing benchmarks but also excels in streaming interaction scenarios. 

\section{Limitations}
While StreamChat demonstrates significant advancements in streaming interaction capabilities for large multimodal models (LMMs), several limitations remain. One limitation is that the timestamps for each text token are generated heuristically from coarse-grained temporal annotation rather than being manually annotated. This reliance on heuristics may introduce inaccuracies in temporal alignment, particularly in complex video scenarios where precise timing is crucial.

{
    \small
    \bibliographystyle{ieeenat_fullname}
    \bibliography{main}
}

\end{document}